%% file: main.tex
\documentclass[lettersize,journal,twoside]{IEEEtran}

\usepackage{amsmath,amsfonts}
\usepackage{algorithmic}
\usepackage{algorithm}
\usepackage{array}
\usepackage[caption=false,font=normalsize,labelfont=sf,textfont=sf]{subfig}
\usepackage{textcomp}
\usepackage{stfloats}
\usepackage{url}
\usepackage{verbatim}
\usepackage{graphicx}
\usepackage{cite}

\usepackage[numbers]{natbib} 
\usepackage{multicol}
\usepackage[bookmarks=true]{hyperref}
\usepackage{amssymb}
\usepackage{amsthm}
\usepackage{amsfonts}
\usepackage{mathtools}
\usepackage{bbm}
\usepackage{tikz}
\usepackage{dsfont}
\usepackage{optidef}
\usepackage{nccmath} 
\usepackage{xcolor}
\usepackage[inkscapelatex=false]{svg}
\usepackage[export]{adjustbox}
\usepackage{comment}
\usepackage[left=16.9mm,right=16.9mm,top=20.1mm,bottom=15.2mm]{geometry}
\usepackage{subcaption}

\begin{document}

\title{Conformal Safety Monitoring for Flight Testing: \\ A Case Study in Data-Driven Safety Learning}

\author{Aaron O. Feldman$^{1}$, D. Isaiah Harp$^{2}$, Joseph Duncan$^{3}$, Mac Schwager$^{1}$, \textit{Senior Member, IEEE}


\thanks{$^{1}$Department of Aeronautics and Astronautics, Stanford University, Stanford, CA 94305, USA, {\texttt\footnotesize \{aofeldma, schwager\}@stanford.edu}. $^{2}$ Department of Aeronautics, United States Air Force Academy, Air Force Academy, CO 80840, USA, daniel.harp@afacademy.af.edu. $^{3}$ %
United States Air Force, Edwards AFB, CA 93523, USA, joseph.duncan.2@us.af.mil.}%
}

\maketitle

\input{Abstract}
\input{Intro}
\input{RelatedWork}
\input{Conformal}

\input{ProblemSetting}
\input{Approach}

\input{Experiments}

\input{Conclusion}


\bibliographystyle{ieeetr} 
\bibliography{refs}

\end{document}

%% file: Abstract.tex
\begin{abstract}
We develop a data-driven approach for runtime safety monitoring in flight testing, where pilots perform maneuvers on aircraft with uncertain parameters. Because safety violations can arise unexpectedly as a result of these uncertainties, pilots need clear, preemptive criteria to abort the maneuver in advance of safety violation. To solve this problem, we use offline stochastic trajectory simulation to learn a calibrated statistical model of the short-term safety risk facing pilots. We use flight testing as a motivating example for data-driven learning/monitoring of safety due to its inherent safety risk, uncertainty, and human-interaction. However, our approach consists of three broadly-applicable components: a model to predict future state from recent observations, a nearest neighbor model to classify the safety of the predicted state, and classifier calibration via conformal prediction. We evaluate our method on a flight dynamics model with uncertain parameters, demonstrating its ability to reliably identify unsafe scenarios, match theoretical guarantees, and outperform baseline approaches in preemptive classification of risk.
\end{abstract}


%% file: Intro.tex
\section{Introduction}
\label{sec: intro}

Safety in human-in-the-loop robotic systems often depends on uncertain dynamics, implicit constraints, and the need for preemptive intervention, factors that make formal specification of safety requirements difficult or incomplete. In this work, we develop a data-driven approach for runtime safety monitoring in flight testing, using conformal prediction to provide statistically calibrated, data-driven safety alerts to test pilots. Our goal is to learn, from prior flight data, a quantitative measure of short-term risk, enabling the system to preemptively signal when a pilot should abort a maneuver to ensure safety. Flight testing presents a compelling case study for reasoning about intangible safety constraints that emerge from uncertainty, human interaction, and implicit domain knowledge. 

Our motivation is grounded in three key challenges. Firstly, flight testing is safety-critical, necessitating principled mechanisms for reasoning about safety. In flight testing, pilots perform aggressive maneuvers on aircraft with partially known dynamics to refine model parameters. These test flights are inherently risky due to the uncertain aircraft behavior, especially near performance limits. Secondly, due to human-machine interaction in flight testing, our monitor must anticipate future safety violation. To be actionable, alerts must allow the human pilot time to react i.e., to abort the maneuver. The need to preemptively alert makes even a well-defined safety specification (e.g., a limit on lateral acceleration) potentially complex/ambiguous; we must reason about the potential of the current state to reach an unsafe future state based on the system dynamics. Thirdly, flight testing requires stochastic analysis of the implicit/future safety constraints. Because the aircraft parameters are uncertain and unknown during the maneuver, we cannot exactly propagate the system dynamics to predict the future state to reason about future safety. Instead, we must reason statistically.

To address these challenges, we propose using simulated flight rollouts to model and calibrate a statistical runtime safety monitor offline. Our method consists of three stages: future state prediction, safety classification for this prediction, and conformal calibration, discussed further in Section~\ref{sec: approach}.

At runtime, our safety monitor operates without requiring forward simulation or explicit evaluation of abstract safety specifications, making it computationally tractable. Using current observations, the predictive model and classifier are queried to obtain a conformal $p$-value, a calibrated scalar measure of risk, enabling both binary alerting and continuous safety scoring. By construction, our approach ensures that the probability of failing to preemptively alert before a true safety violation is at most a user-specified $\epsilon$, offering a statistical guarantee that is easy to interpret for the human operator/pilot. 

This work thus illustrates an approach to provide data-driven operational safety guarantees in settings like flight testing, where uncertain dynamics and human-machine interaction can make traditional safety constraint specification challenging.

%% file: RelatedWork.tex
\section{Literature Review}
\label{sec: lit_rev}

Some related work in the field of flight testing includes \cite{MorelliModel} and \cite{SafeEnvelope}. \cite{MorelliModel} fits a model to characterize plane parameters (e.g., lift, drag, and moment coefficient) across a variety of aircraft. This work demonstrates a procedure for data-driven aircraft parameter estimation and uncertainty quantification rather than tackling the problem of runtime safety monitoring. In fact, the resulting parameter bounds and polynomial models from \cite{MorelliModel} could be used for stochastic simulation within our approach as a next step. \cite{SafeEnvelope} presents several approaches for real-time runtime safety monitoring of flight tests. One of their approaches is to determine a safe operating domain based on trajectories flown in simulation with some plane parameter randomization. They also consider an online approach for recursively estimating the aircraft stability/control derivatives and abort if these fall outside user-defined bounds. In contrast, our approach implicitly defines a safe operating domain using a nearest neighbor classifier and, by predicting future states using a buffer of recent data, implicitly conditions on aircraft parameters without explicitly regressing them.

Outside of flight testing, conformal prediction has been widely used for trajectory prediction in robotics (e.g., for pedestrians) \cite{lindemann2023safe, cleaveland2023conformal, strawn2023conformal, dixit2022adaptive, muthali2023multiagent, shape_template_cp}, sensor anomaly detection \cite{sinha2023systemlevelviewoutofdistributiondata, laxhammar2010conformal, laxhammar2011, smith2014anomaly, contreras2024outofdistributionruntimeadaptationconformalized, sinha2023closingloopruntimemonitors}, and dynamics model error \cite{conformal_sys_error}.

The work most closely related to this one is our own prior research \cite{feldman2025learningrobotsafetysparse} which used conformal prediction to learn a set of unsafe states from human feedback. This project builds on this work in two regards. Firstly, we apply the theory of \cite{feldman2025learningrobotsafetysparse} to the novel and practical use case of flight testing. Secondly, we combine the nearest neighbor model with a predictive model to predict future outputs. We show that this predictive model is key to the success of our approach and outperforms other transformations like PCA that were used in \cite{feldman2025learningrobotsafetysparse}.

%% file: Conformal.tex
\section{Overview of Conformal Prediction}
\label{sec: cp_overview}
Conformal prediction \cite{shafer2007tutorial, angelopoulos2022gentle} is a statistical approach for uncertainty quantification that operates without distributional assumptions. The core result is that for $N+1$ random variables (scores) $s_1,...,s_N, s_{N+1}$ exchangeable, equally likely under re-ordering,
\begin{equation}
\label{eq: marginal}
\Pr(s_{N+1} \leq s_{(k)}) \geq \frac{k}{N+1}
\end{equation}
with equality assuming no ties. Thus, for a user-specified $\epsilon$, taking $k(\epsilon) = \lceil(N+1)(1-\epsilon)\rceil$ will ensure that 
\begin{equation}
\Pr(s_{N+1} \leq s_{(k)}) \geq 1-\epsilon
\end{equation}
i.e., we will fail to bound $s_{N+1}$ with a miss rate of at most $\epsilon$. Here, $s_{(k)}$ refers to the $k$-th order statistic of $s_1,...,s_N$. Broadly, there are two categories of conformal prediction using these results: split conformal prediction \cite{vovkinductiveCP} and full conformal prediction \cite{vovktransductiveCP}. We build on \cite{feldman2025learningrobotsafetysparse}, which used full conformal prediction within a nearest neighbor model. Using conformal prediction, we can assign to the nearest neighbor model's output a statistical interpretation and determine when to alert to achieve a user-specified miss rate $\epsilon$.


%% file: ProblemSetting.tex
\section{Problem Setting}
\label{sec: problem_setting}

We assume a distribution over unknown system (plane) parameters $\theta \sim D_{\theta}$ from which we can repeatedly sample $\theta_i \sim D_{\theta}$ in simulation to produce resulting trajectories $\tau_i$. The true, unknown, plane parameters are viewed as a new draw $\theta^* \sim D_{\theta}$ and induce a deterministic test trajectory $\tau^*$. Our goal is to develop an early warning system such that if $\tau^*$ will become unsafe, we will alert the pilot with $t_{early}$ advanced notice (we use $t_{early} = 0.25$ [sec]), leaving time for the human to take corrective action.

As an illustrative example for our experiments, we consider a continuous, linear time invariant flight dynamics model, provided to us by members of the USAF test pilot school.
\begin{equation}
\dot{x} = A x + B u, y = C x + D u.
\end{equation}
The three-dimensional plane state $x = (\beta, p, r)$ consists of the sideslip angle ([rad]), roll rate ([rad/s]), and yaw rate ([rad/s]) respectively. The control action $u = (\delta_a, \delta_r)$ consists of the aileron and rudder deflection ([rad]). The output $y = (\beta, p, r, N_y, \delta_a, \delta_r)$ stores the state, action, and additionally $N_y$ the lateral acceleration ([g]). The matrices $\theta = (A, B, C, D)$ are the unknown system parameters in our setting and are randomized over a known range (e.g., based on prior flight data). On top of the open-loop transfer function $G(s) = C (s I - A)^{-1} B + D$, a feedback controller $K(s)$ is modeled to translate pilot inputs into control actions, and we simulate rollouts using the closed-loop transfer function. We consider a simple flight maneuver referred to as a rudder doublet wherein the pilot applies as input to the closed-loop system $\delta_r = 1$ followed by $\delta_r = -1$ for one second each.

We consider that a ``high-level" abstract safety specification can be queried to evaluate each rollout's safety in hindsight offline. For instance, this might involve performing a computationally expensive structural/load analysis at different times during the maneuver or an evaluation of the changing gain/phase margin for a nonlinear dynamical system. This safety analysis could even come from simulated pilot intervention/interruption in hindsight \cite{feldman2025learningrobotsafetysparse}. For preliminary analysis in this work, a rollout is defined to become unsafe if the lateral acceleration grows too large: $|N_y| \geq 0.5$. Even in this case where the specification is well-defined at the current time, preemptively reasoning about violation remains challenging due to uncertainty in the true aircraft parameters.


%% file: Approach.tex
\section{Approach}
\label{sec: approach}

\subsection{Data Collection}

Offline, the user specifies a number $N$ of unsafe trajectories to collect, and we repeatedly sample system parameters $\theta_i$ and perform rollouts $\tau_i$ until we have obtained $N$ unsafe trajectories $D_u$. In this process, we also obtain a variable number, say $M$, of safe trajectories $D_s$. 

Each raw trajectory consists of a sequence of outputs $\tau = (y_1, ..., y_T)$. To preemptively predict system failure, reasoning about the single, current output $y_t$ is typically insufficient to anticipate future behavior. To form a better feature we concatenate a short buffer of the recent outputs $o_t = (y_{t-k}, ..., y_t)$ as our observation at each time.

Additionally, for each unsafe trajectory $\tau_i \in D_u$, we record the time of failure $t_i$ and go back in time $t_{early}$ to obtain the observation $o_i := o_{t_i - t_{early}}$. We collect the observations $o_i, \ i \in \{1,...,N\}$, taken $t_{early}$ in advance of system failure, into $\mathcal{O}_u$, which we refer to as the error observation set. For each safe trajectory $\tau_i \in D_s$, we extract observations, possibly randomly subselecting a few times from each trajectory, to construct a similar set $\mathcal{O}_s$ of safe observations. We then offline fit and calibrate a classifier to distinguish between $\mathcal{O}_u$ and $\mathcal{O}_s$.

Our approach consists of three components:
\begin{itemize}
    \item State Prediction: A model is trained to forecast future system outputs from a short history of observed states.
    \item Safety Classification: A nearest-neighbor classifier identifies if the predicted future state is likely safe or unsafe.
    \item Conformal Calibration: Using conformal prediction, we calibrate the classifier's alert threshold to guarantee that the miss rate does not exceed a user-defined bound.
\end{itemize}

\subsection{State Prediction}

We use a linear model to predict the state $t_{early}$ into the future from the current observation:
\begin{equation}
    \hat{y}_{t + t_{early}} = \phi(o_t) = M o_t + \mu
\end{equation}
where $M, \mu$ are learned parameters of the linear model $\phi$. We fit this model using least squares to observations from the safe trajectories $D = \{(o_t, y_{t+t_{early}})\}$ using the future state $y_{t+t_{early}}$ as the regression target to predict i.e., we solve
\begin{equation}
    \min_{M, \mu} \sum_{(o_t, y_{t + t_{early}}) \in D} ||y_{t + t_{early}} - (M o_t + \mu)||_2^2.
\end{equation}
Using the resulting linear model, we transform the observations in $\mathcal{O}_u, \mathcal{O}_s$ to their predicted future states $\hat{y}_{t + t_{early}}$, obtaining corresponding sets $\mathcal{Y}_u = \phi(\mathcal{O}_u), \mathcal{Y}_s = \phi(\mathcal{O}_s)$.

We could instead directly apply nearest neighbor classification to distinguish between $\mathcal{O}_u, \mathcal{O}_s$. However, by first transforming to the predicted future states and classifying in the transformed space of $\mathcal{Y}_u, \mathcal{Y}_s$ we can improve the classifier performance, as shown in our experiments. The linear model can be viewed as learning a particularly useful representation for downstream classification.

\subsection{Safety Classification}
After transforming to the predicted future state $\hat{y}_{t + t_{early}}$ we could simply check the safety specification for it. However, this may be impossible at runtime if the safety specification requires significant computation, privileged information, or expert/pilot labeling. Even if we can check the safety specification at runtime, this would be imperfect as we have only a predicted, not true, future state.

Therefore, we further fit a nearest neighbor classifier to distinguish $\mathcal{Y}_u$ from $\mathcal{Y}_s$. We use the conformal score from \cite{feldman2025learningrobotsafetysparse} which scores a test point $y$ by its distance to the nearest unsafe versus safe point in the data:
\begin{equation}
\label{eq: nn_score}
    s(y) = \min_{y_u \in \mathcal{Y}_u} ||y - y_u||_2^2 - \min_{y_s \in \mathcal{Y}_s} ||y - y_s||_2^2.
\end{equation}
Thus, we expect new unsafe points to have low scores.

\subsection{Conformal Prediction Calibration}
Lastly, we calibrate our nearest neighbor classifier using conformal prediction. Given a new test point $y$, we would like to give a statistical interpretation to the score $s(y)$ output from the nearest neighbor model. Practically, we need to determine a threshold at which $s(y)$ is declared too low so that our warning system should issue an alert. We set this alert threshold to ensure a user-specified miss rate $\epsilon$ i.e., a requirement that the warning system miss preemptively flagging ($t_{early}$ in advance) in only $\epsilon$ fraction of trajectories which became unsafe. 

\cite{feldman2025learningrobotsafetysparse} showed that we could calibrate the nearest neighbor classifier by simply recording the intra-dataset nearest neighbor distances, which for unsafe observation $y_u^i \in \mathcal{Y}_u$ is given by
\begin{equation}
    \alpha_i = \min_{y_u' \in \mathcal{Y}_u - \{y_u^i\}} - \min_{y_s \in \mathcal{Y}_s} ||y_u^i - y_s||_2^2.
\end{equation}
Notably, this allows us to calibrate offline, reusing $\mathcal{Y}_u$ without need for a separate validation/calibration dataset. To achieve a miss rate at most $\epsilon$, the conformal score threshold $s^*$ is
\begin{equation}
\label{eq: nn_cutoff}
    s^* = \alpha_{(k)}, k = \lceil(N+1)(1-\epsilon)\rceil.
\end{equation}
where $\alpha_{(k)}$ refers to the $k$-th order statistic of $\alpha_1, ..., \alpha_N$. If at test time $s(y) \leq s^*$, we issue an alert.

Beyond a single threshold dictating when we should alert, we can obtain the conformal $p$-value associated with $s(y)$ as a preemptive measure of the risk facing the pilot. For given $s(y)$, the associated $p$-value $\epsilon^*$ is defined as the smallest $\epsilon$ for which we would not alert based on Eq.~\ref{eq: nn_cutoff} \footnote{This can be found by determining the index $k$ at which $s(y)$ should be inserted into a sorted list of $\alpha_1, ..., \alpha_N$ and then mapping to an $\epsilon$ via Eq.~\ref{eq: nn_cutoff}}. Practically, the $p$-value ranges from $[0,1]$ with lower values deemed safer, and alerting whenever $\epsilon^* \geq \epsilon$ is equivalent to alerting whenever $s(y) \leq s^*$. Thus, the pilot can simply monitor the $p$-value during flight and abort if it exceeds designated miss rate $\epsilon$.

\subsection{Runtime Safety Monitoring}

During deployment, we perform the following at each time:
\begin{enumerate}
    \item Form observation $o_t = (y_{t-k}, ..., y_t)$ using latest outputs.
    \item Predict future state $\hat{y}_{t + t_{early}} = \phi(o_t)$.
    \item Compute nearest neighbor score $s(\hat{y}_{t + t_{early}})$ as in Eq.~\ref{eq: nn_score}.
    \item Convert this score to the associated conformal $p$-value.
    \item Report the $p$-value to the pilot and alert if it exceeds user-specified miss rate $\epsilon$.
\end{enumerate}

%% file: Experiments.tex
\section{Experiments}
\label{sec: experiments}

In our experiments, we used the system described in Section~\ref{sec: problem_setting} and collected $N = 50$ unsafe trajectory rollouts in simulation with randomized plane parameters and discretization $\Delta t = 0.05$ [sec]. We specify that our warning system should alert $t_{early} = 0.25$ [sec] in advance of failure. For state prediction, we use a delay of $3$ states i.e., $o_t = (y_{t-2}, y_{t-1}, y_t)$. Using the process described in Section~\ref{sec: approach}, we subsampled $50$ observations per safe trajectory to fit the linear predictive model and for the nearest neighbor classification.

\begin{figure}
    \centering
    \includegraphics[width=0.9\linewidth]{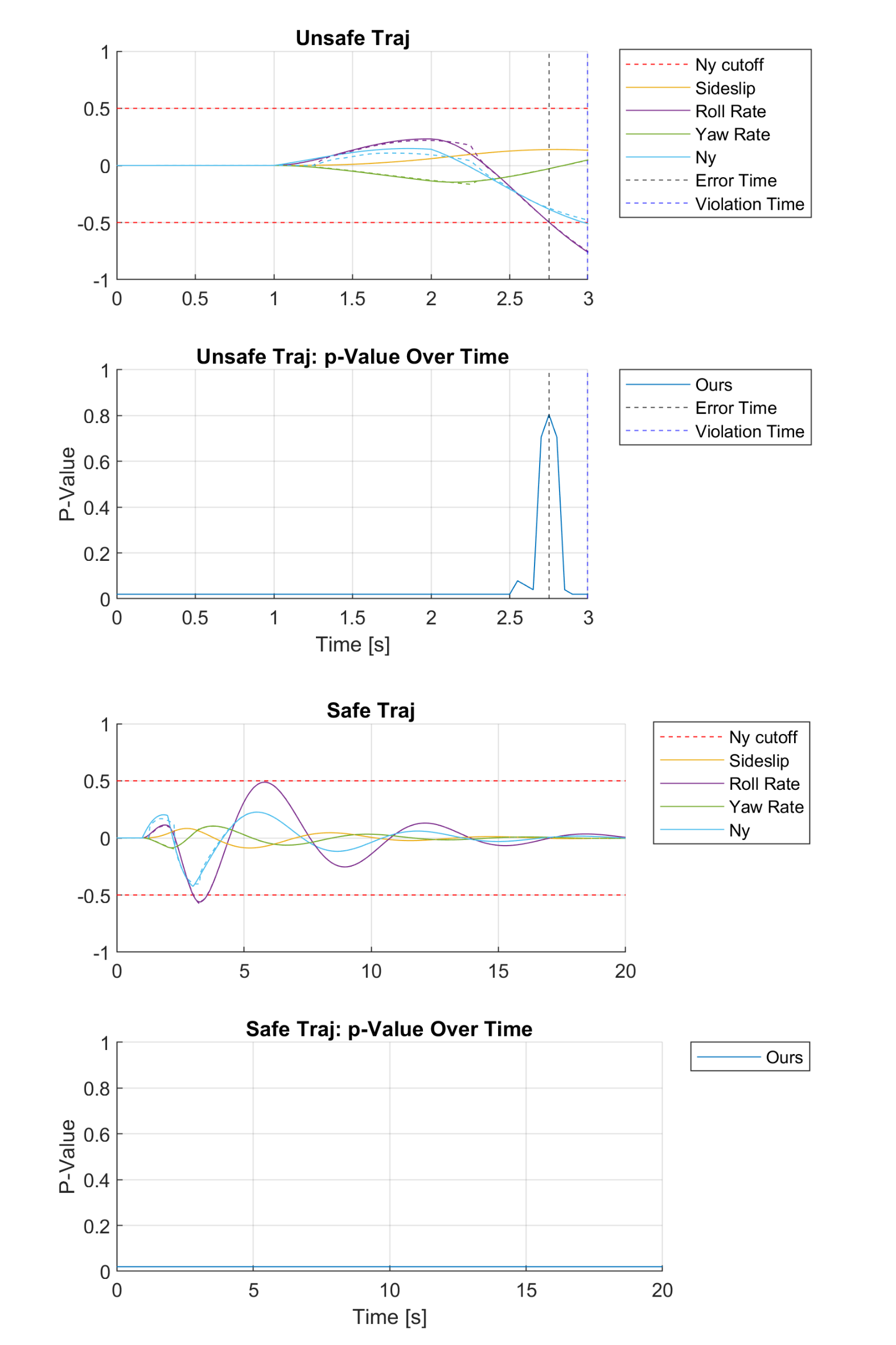}
    \caption{True versus Predicted Outputs for Unsafe/Safe Trajectories with Associated $p$-Values}
    \label{fig: merged}
\end{figure}

In Figure~\ref{fig: merged} we show an unsafe and safe trajectory for two test rollouts of the rudder doublet maneuver. For the unsafe trajectory, we cut simulation at the time of system failure, marked with a dashed black line. We show the time $t_{early}$ before then when the alert should trigger, as a blue dashed line. The safety cutoff of $|N_y| = 0.5$ is shown with horizontal red lines. We overlay the predicted outputs $\hat{y}_{t + t_{early}}$ with dashed lines of the same color as the true outputs. All predicted components match ground-truth well for both trajectories. Below each unsafe/safe trajectory, we show the associated $p$-value over time. Our runtime monitor performs well as the $p$-value rises near the time of system failure in the unsafe case, peaking around $t_{early}$ before failure. In contrast, it remains at the lowest value during the safe trajectory.

To evaluate our method systematically, we varied the target miss rate $\epsilon$ and measured empirical miss rate and classification power. We trained the monitor on 10 different datasets, with $N = 50$ unsafe rollouts each, and tested on 500 trajectories.

As alternatives to our approach we considered
\begin{itemize}
    \item ``No pred'': drops prediction model $\phi$, applying conformal nearest neighbor directly to observation $o_t$.
    \item ``PCA": replaces $\phi$ with PCA to reduce $o_t$ dimension (from 18 to 6).
    \item ``Current $|N_y|$": replaces the conformal score with $s(y) = -|N_y|$ which increases as the safety limit is approached.
    \item ``Predicted $|N_y|$": applies $s(y) = -|N_y|$ to the predicted future state $\hat{y}_{t+t_{early}}$.
\end{itemize}

\begin{figure}
    \centering
    \includegraphics[width=0.9\linewidth]{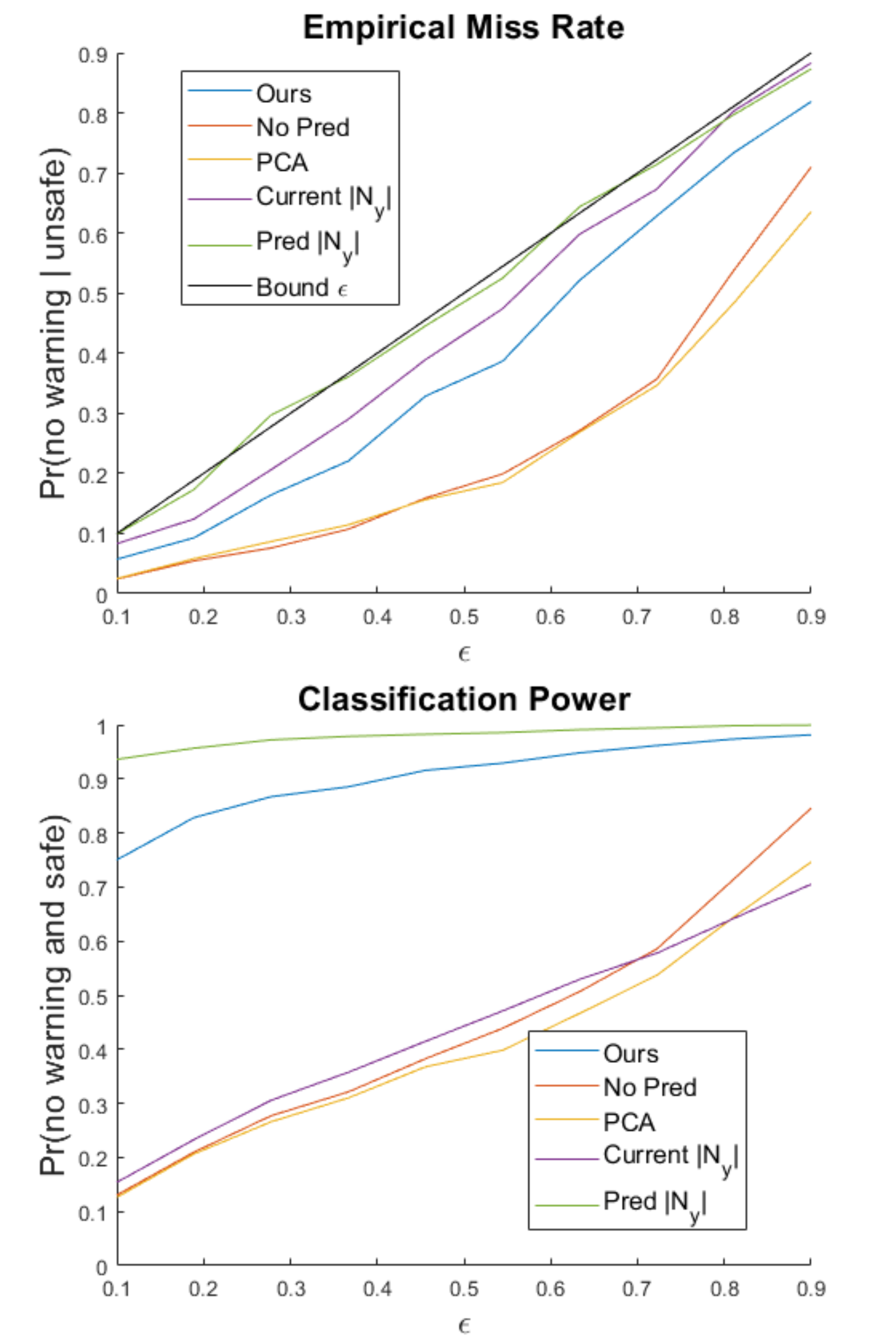}
    \caption{Miss Rate and Classification Power for Varying $\epsilon$}
    \label{fig: systematic}
\end{figure}

The upper plot of Figure~\ref{fig: systematic} shows the empirical miss rate, the fraction of unsafe test trajectories where no alert is issued by $t_{early}$ seconds before failure. The empirical miss rate falls below the theoretical upper bound of $\epsilon$, validating the conformal guarantee. For ``Pred $|N_y|$'' the bound appears slightly violated, due to taking an empirical average, but this would vanish with more fits. The bottom plot of Figure~\ref{fig: systematic} shows the classification power i.e., the probability of no alert during a safe trajectory, which should ideally remain high across $\epsilon$.

Our approach retains good classification power across the $\epsilon$ range and outperforms all alternatives except ``Pred $|N_y|$''. Unlike our general-purpose nearest neighbor score, this baseline uses the known safety specification $|N_y| \leq 0.5$ to handcraft a conformal score. For more complex or implicit safety specifications, such tailored scores may be unavailable or impractical. Still, the success of ``Pred $|N_y|$'' over ``Current $|N_y|$'' highlights the value of the prediction model. Similarly, we note that replacing the prediction model with PCA, designed for data reconstruction not prediction, performs poorly.

%% file: Conclusion.tex
\section{Conclusion}
\label{sec: conclusion}

This work serves as a proof-of-concept for providing test pilots with a statistics-based preemptive runtime safety monitor. Our approach, featuring a prediction model, nearest neighbor classification, and conformal prediction, could be used more generally for data-driven safety learning.

There are several exciting directions for future research. It would be valuable to use stochastic and reactive human models within the runtime monitor. The current prediction step could then also forecast the human's actions, in turn changing the predicted future state. Secondly, it would be valuable to test with more abstract/complex safety specifications (e.g., based on structural/load analysis, gain/phase margin, or pilot intervention) where a custom conformal score cannot be easily defined. Thirdly, it would be interesting to consider other prediction models, which could be probabilistic or adapt online.

%% file: refs.bib
@article{SafeEnvelope,
title = {Enhancing Real Time Monitoring Support for Safe Envelope Expansion},
journal = {IFAC-PapersOnLine},
volume = {49},
number = {1},
pages = {254-259},
year = {2016},
note = {4th IFAC Conference on Advances in Control and Optimization of Dynamical Systems ACODS 2016},
issn = {2405-8963},
doi = {https://doi.org/10.1016/j.ifacol.2016.03.062},
url = {https://www.sciencedirect.com/science/article/pii/S2405896316300623},
author = {Khadeeja {Nusrath TK} and Dushyant Kaliyari and Jatinder Singh and Vijay V Patel}
}

@article{MorelliModel,
author = {Grauer, Jared A. and Morelli, Eugene A.},
title = {Generic Global Aerodynamic Model for Aircraft},
journal = {Journal of Aircraft},
volume = {52},
number = {1},
pages = {13-20},
year = {2015},
doi = {10.2514/1.C032888},
URL = {https://doi.org/10.2514/1.C032888},
eprint = {https://doi.org/10.2514/1.C032888}
}

@misc{shape_template_cp,
      title={Multi-Modal Conformal Prediction Regions with Simple Structures by Optimizing Convex Shape Templates}, 
      author={Renukanandan Tumu and Matthew Cleaveland and Rahul Mangharam and George J. Pappas and Lars Lindemann},
      year={2024},
      eprint={2312.07434},
      archivePrefix={arXiv},
      primaryClass={cs.LG},
      url={https://arxiv.org/abs/2312.07434}, 
}

@misc{feldman2025learningrobotsafetysparse,
      title={Learning Robot Safety from Sparse Human Feedback using Conformal Prediction}, 
      author={Aaron O. Feldman and Joseph A. Vincent and Maximilian Adang and Jun En Low and Mac Schwager},
      year={2025},
      eprint={2501.04823},
      archivePrefix={arXiv},
      primaryClass={cs.RO},
      url={https://arxiv.org/abs/2501.04823}, 
}

@inproceedings{vovktransductiveCP,
author = {Gammerman, A. and Vovk, V. and Vapnik, V.},
title = {Learning by transduction},
year = {1998},
isbn = {155860555X},
publisher = {Morgan Kaufmann Publishers Inc.},
address = {San Francisco, CA, USA},
booktitle = {Proceedings of the Fourteenth Conference on Uncertainty in Artificial Intelligence},
pages = {148–155},
numpages = {8},
location = {Madison, Wisconsin},
series = {UAI'98}
}

@inproceedings{vovkinductiveCP,
  title={Inductive confidence machines for regression},
  author={Papadopoulos, Harris and Proedrou, Kostas and Vovk, Volodya and Gammerman, Alex},
  booktitle={Machine learning: ECML 2002: 13th European conference on machine learning Helsinki, Finland, August 19--23, 2002 proceedings 13},
  pages={345--356},
  year={2002},
  organization={Springer}
}

@article{shafer2007tutorial,
  title={A tutorial on conformal prediction.},
  author={Shafer, Glenn and Vovk, Vladimir},
  journal={Journal of Machine Learning Research},
  volume={9},
  number={3},
  year={2008}
}

@article{angelopoulos2022gentle,
  title={A gentle introduction to conformal prediction and distribution-free uncertainty quantification},
  author={Angelopoulos, Anastasios N and Bates, Stephen},
  journal={arXiv preprint arXiv:2107.07511},
  year={2021}
}

@InProceedings{conformal_sys_error,
  title = 	 {Uncertainty quantification and robustification of model-based controllers using conformal prediction},
  author =       {Chee, Kong Yao and Silva, Thales C. and Hsieh, M. Ani and Pappas, George J.},
  booktitle = 	 {Proceedings of the 6th Annual Learning for Dynamics and Control Conference},
  pages = 	 {528--540},
  year = 	 {2024},
  editor = 	 {Abate, Alessandro and Cannon, Mark and Margellos, Kostas and Papachristodoulou, Antonis},
  volume = 	 {242},
  series = 	 {Proceedings of Machine Learning Research},
  month = 	 {15--17 Jul},
  publisher =    {PMLR},
  url = 	 {https://proceedings.mlr.press/v242/chee24a.html}
}

@article{sinha2023systemlevelviewoutofdistributiondata,
  title={A system-level view on out-of-distribution data in robotics},
  author={Sinha, Rohan and Sharma, Apoorva and Banerjee, Somrita and Lew, Thomas and Luo, Rachel and Richards, Spencer M and Sun, Yixiao and Schmerling, Edward and Pavone, Marco},
  journal={arXiv preprint arXiv:2212.14020},
  year={2022}
}

@inproceedings{laxhammar2010conformal,
  title={Conformal prediction for distribution-independent anomaly detection in streaming vessel data},
  author={Laxhammar, Rikard and Falkman, G{\"o}ran},
  booktitle={Proceedings of the first international workshop on novel data stream pattern mining techniques},
  pages={47--55},
  year={2010}
}

@inproceedings{laxhammar2011,
  title={Sequential conformal anomaly detection in trajectories based on hausdorff distance},
  author={Laxhammar, Rikard and Falkman, G{\"o}ran},
  booktitle={14th international conference on information fusion},
  pages={1--8},
  year={2011},
  organization={IEEE}
}

@inproceedings{smith2014anomaly,
  title={Anomaly detection of trajectories with kernel density estimation by conformal prediction},
  author={Smith, James and Nouretdinov, Ilia and Craddock, Rachel and Offer, Charles and Gammerman, Alexander},
  booktitle={Artificial Intelligence Applications and Innovations: AIAI 2014 Workshops: CoPA, MHDW, IIVC, and MT4BD, Rhodes, Greece, September 19-21, 2014. Proceedings 10},
  pages={271--280},
  year={2014},
  organization={Springer}
}

@article{contreras2024outofdistributionruntimeadaptationconformalized,
  title={Out-of-Distribution Runtime Adaptation with Conformalized Neural Network Ensembles},
  author={Contreras, Polo and Shorinwa, Ola and Schwager, Mac},
  journal={arXiv preprint arXiv:2406.02436},
  year={2024}
}

@inproceedings{sinha2023closingloopruntimemonitors,
  title={Closing the loop on runtime monitors with Fallback-Safe MPC},
  author={Sinha, Rohan and Schmerling, Edward and Pavone, Marco},
  booktitle={2023 62nd IEEE Conference on Decision and Control (CDC)},
  pages={6533--6540},
  year={2023},
  organization={IEEE}
}

@article{lindemann2023safe,
  title={Safe planning in dynamic environments using conformal prediction},
  author={Lindemann, Lars and Cleaveland, Matthew and Shim, Gihyun and Pappas, George J},
  journal={IEEE Robotics and Automation Letters},
  year={2023},
  publisher={IEEE}
}

@inproceedings{cleaveland2023conformal,
  title={Conformal prediction regions for time series using linear complementarity programming},
  author={Cleaveland, Matthew and Lee, Insup and Pappas, George J and Lindemann, Lars},
  booktitle={Proceedings of the AAAI Conference on Artificial Intelligence},
  volume={38},
  number={19},
  pages={20984--20992},
  year={2024}
}

@article{strawn2023conformal,
  title={Conformal predictive safety filter for rl controllers in dynamic environments},
  author={Strawn, Kegan J and Ayanian, Nora and Lindemann, Lars},
  journal={IEEE Robotics and Automation Letters},
  year={2023},
  publisher={IEEE}
}

@inproceedings{dixit2022adaptive,
  title={Adaptive conformal prediction for motion planning among dynamic agents},
  author={Dixit, Anushri and Lindemann, Lars and Wei, Skylar X and Cleaveland, Matthew and Pappas, George J and Burdick, Joel W},
  booktitle={Learning for Dynamics and Control Conference},
  pages={300--314},
  year={2023},
  organization={PMLR}
}

@inproceedings{muthali2023multiagent,
  title={Multi-agent reachability calibration with conformal prediction},
  author={Muthali, Anish and Shen, Haotian and Deglurkar, Sampada and Lim, Michael H and Roelofs, Rebecca and Faust, Aleksandra and Tomlin, Claire},
  booktitle={2023 62nd IEEE Conference on Decision and Control (CDC)},
  pages={6596--6603},
  year={2023},
  organization={IEEE}
}
